\documentclass[11pt]{article}

\usepackage[preprint]{acl}

\usepackage{times}
\usepackage{latexsym}

\usepackage[T1]{fontenc}

\usepackage[utf8]{inputenc}

\usepackage{microtype}

\usepackage{inconsolata}

\usepackage{graphicx}

\usepackage[utf8]{inputenc} 
\usepackage[T1]{fontenc}    
\usepackage{hyperref}       
\usepackage{url}            
\usepackage{booktabs}       
\usepackage{amsfonts}       
\usepackage{nicefrac}       
\usepackage{microtype}      
\usepackage[dvipsnames]{xcolor}         
\usepackage{graphicx} 
\usepackage{parskip}  
\usepackage{setspace} 
\usepackage{refcount} 
\usepackage{upquote}  
\usepackage{fancyhdr}
\usepackage{lastpage}       
\usepackage{lscape}         
\usepackage{verbatim}       
\usepackage{listings}       
\usepackage{epsfig}         
\usepackage{array}          
\usepackage{enumitem}       
\usepackage{hhline}         
\usepackage{siunitx}        
\usepackage{amsmath}        
\usepackage{amssymb}        
\usepackage{amsthm}         
\usepackage{algorithm, algorithmicx, algpseudocode}
\usepackage{ifthen}         
\usepackage{float}
\newfloat{algorithm}{t}{lop}
\usepackage{cleveref}
\usepackage{booktabs}
\usepackage{makecell}

\usepackage{pdfpages}
\usepackage{subcaption}
\usepackage{multirow}
\usepackage{multicol}
\usepackage{wrapfig}

\algrenewcommand\algorithmicindent{0.8em}%
\title{Deep Kernel Fusion for Transformers}

\author{Zixi Zhang \\
  Imperial College London \\
  London, UK \\
  \texttt{b.zhang25@imperial.ac.uk} \\\And
  Zhiwen Mo \\
  Imperial College London\\
  London, UK \\
  \texttt{zhiwen.mo25@imperial.ac.uk} \\\AND
  Yiren Zhao \\
  Imperial College London\\
  London, UK \\
  \texttt{a.zhao@imperial.ac.uk} \\\And
  Robert Mullins \\
  University of Cambridge\\
  Cambridge, UK \\
  \texttt{robert.mullins@cl.cam.ac.uk}
}

\begin{document}
\maketitle
\begin{abstract}
  Agentic LLM inference with long contexts is increasingly limited by memory bandwidth rather than compute. In this setting, SwiGLU MLP blocks, whose large weights exceed cache capacity, become a major yet under-optimized bottleneck. We propose \textbf{DeepFusionKernel}, a deeply fused kernel that cuts HBM traffic and boosts cache reuse, delivering up to 13.2\% speedup on H100 and 9.7\% on A100 over SGLang. Integrated with SGLang and paired with a kernel scheduler, DeepFusionKernel ensures consistent accelerations over generation lengths, while remaining adaptable to diverse models, inference configurations, and hardware platforms.
\end{abstract}

\section{Introduction}
\label{sec:intro}

The Transformer architecture~\citep{transformer} underpins modern large language models (LLMs)~\citep{gpt3,gpto1,deepseekr1}, and as these models are deployed in agentic, real-world workloads, inference efficiency becomes a first-order concern~\citep{deepspeed,efficientinferencesurvey,ReAct}. Agentic workloads require the model to handle very long contexts, maintain many persistent state items, and often produce long-form outputs (for example, large codebases). As a result, they process on the order of $100\times$ more tokens per inference than typical chatbot-style workloads~\citep{wu2025combatingmemorywallsoptimization}. In this regime, caching and fast access to KV values shift the primary bottleneck from raw compute to memory capacity and memory bandwidth. That shift in turn changes the optimization landscape: permissible batch sizes shrink, GEMM inputs become ``fat,'' Tensor Core utilization drops, and available compute is severely underused.

Most recent GPU optimizations focus on attention~\citep{flashattn3,vllm,sglang,flashinfer}, but in autoregressive decoding, the SwiGLU MLP blocks~\citep{gatedtransformer,llama3,qwen,deepseekv3} dominate parameter count and drive memory-bandwidth pressure. Although modern accelerators offer high HBM bandwidth~\citep{a100,h100,b200}, memory scaling has not kept pace with compute, so memory-bound kernels remain the limiting factor for throughput on many real deployments.

To close this gap, we propose \textbf{aggressive deep kernel fusion} to reduce memory traffic and improve utilization in FFN blocks in the attention mechanism. We implement a highly optimized fused operator, the \textbf{DeepFusionKernel}, which combines the separate GEMMs and pointwise kernels used in common four-kernel and two-kernel SwiGLU implementations (e.g., PyTorch~\citep{pytorch}, SGLang~\citep{sglang}, and vLLM~\citep{vllm}) into a single fused kernel that minimizes intermediate memory reads/writes and exposes more efficient work for Tensor Cores. Integrated with a lightweight, profile-driven kernel scheduler and deployed inside SGLang, DeepFusionKernel yields consistent, deployable speedups in bandwidth-bound agentic scenarios: up to 9.7\% on A100 and 13.2\% on H100 clusters compared to the SOTA SGLang implementation. 

\begin{figure*}
\centering
  \includegraphics[width=\linewidth]{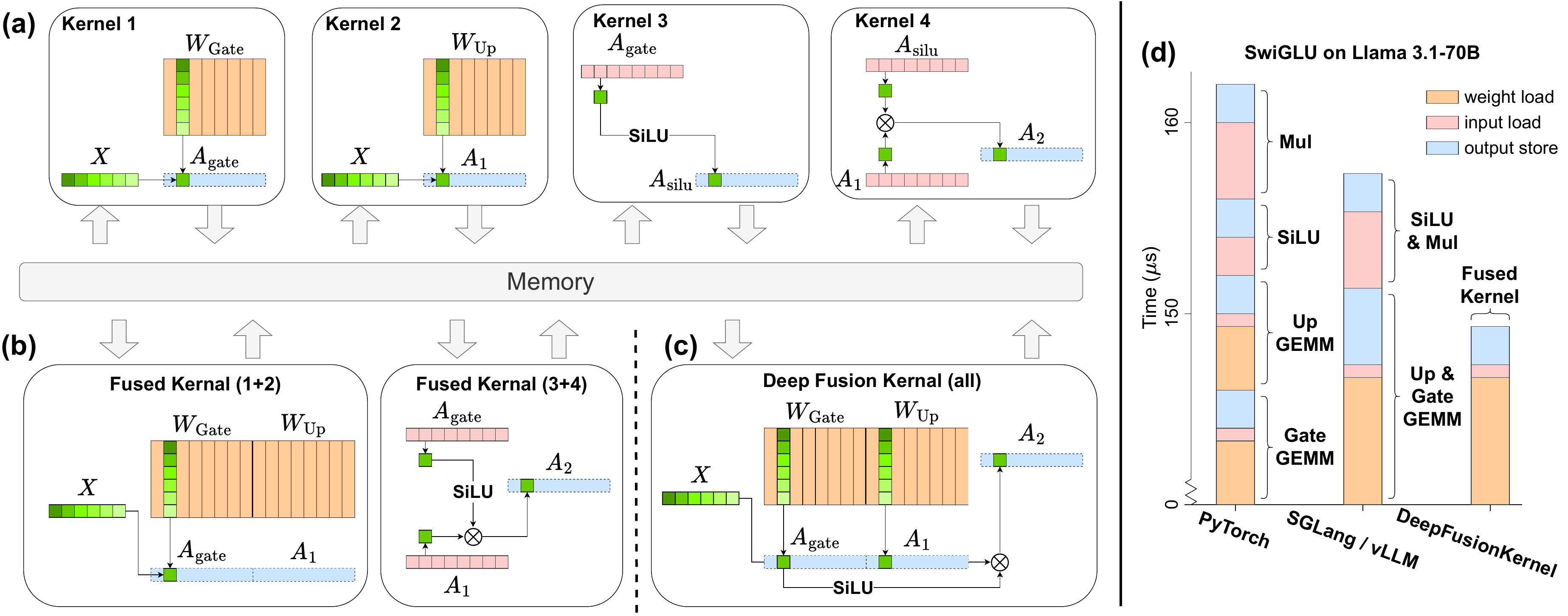} 
\caption{DeepFusionKernel leverages aggressive kernel fusion on SwiGLU blocks to eliminate intermediate activations and reduce memory traffic--without increasing FLOPs. Panels show implementation layouts: (a) naive PyTorch with four kernel launches; (b) the two-kernel design used by SGLang and vLLM; and (c) our single, deeply fused kernel that streams data through GEMMs and nonlinearities to avoid extra loads/stores. Speedup over PyTorch and SGLang / vLLM is displayed in (d). By removing these redundant reads/writes, DeepFusionKernel yields up to 9.7\% and 13.2\% throughput improvements on A100 and H100 GPUs, respectively, on bandwidth-bound autoregressive decoding workloads.}
\label{fig:deepfusionkernel-vs-sglang}
\end{figure*}

\section{Background}

Modern Transformer-based LLMs replace the original two-layer ReLU design in the MLP with the gating-integrated SwiGLU variant~\citep{gatedtransformer}:
$$
\resizebox{.9\hsize}{!}{$
\begin{aligned}
&A_{\rm gate} = XW_{\rm Gate}, \quad A_1 = XW_{\rm Up},\\
&A_{\rm silu} = \mathrm{SiLU}(A_{\rm gate}) = \mathrm{Sigmoid}(A_{\rm gate}) \otimes A_{\rm gate},\\
&A_2 = A_1 \otimes A_{\rm silu},\quad
Y = A_2 W_{\rm Down}.
\end{aligned}
$}
$$
Here $W_\mathrm{Up}, W_\mathrm{Gate} \in \mathbb{R}^{d_\mathrm{model} \times d_\mathrm{ff}}$ and $W_\mathrm{Down} \in \mathbb{R}^{d_\mathrm{ff} \times d_\mathrm{model}}$, with $d_\mathrm{ff}$ being 3.5 to 4 times the $d_\mathrm{model}$. These large matrices dominate the model’s parameter count and memory footprint.


\section{Method}

Kernel fusion~\citep{deepspeed,trt-llm,Apex} reduces end-to-end latency and memory traffic by combining sequences of operations (GEMMs, pointwise nonlinearities, and simple elementwise math) into a single GPU kernel, thereby eliminating intermediate reads and writes--the dominant cost in bandwidth-bound LLM decoding~\citep{vllm,sglang}. Fusion is most effective for elementwise and tile-local computations; true reductions (e.g., Softmax) introduce long-range dependencies that limit cross-SM streaming and therefore are not good fusion targets. Guided by this observation, we split the Transformer MLP into two stages:
\begin{equation*}
\resizebox{.9\hsize}{!}{$A_2 = (XW_\mathrm{Up}) \otimes \mathrm{SiLU}(XW_\mathrm{Gate}) , \quad Y = A_2W_\mathrm{Down}.$}
\end{equation*}
and focus fusion effort on the first stage, which contains the GEMMs and pointwise gating that dominate memory traffic during autoregressive decoding.

As illustrated in \Cref{fig:deepfusionkernel-vs-sglang}, DeepFusionKernel fuses the separate GEMMs and pointwise kernels (four launches in a naive PyTorch implementation, two in SGLang/vLLM) into a single deeply fused operator. The fused kernel streams intermediate values through the computation, avoiding materialization of large temporaries and drastically cutting reads/writes to HBM. To realize this in practice, we systematically explore loop ordering and tiling strategies that maximize on-chip reuse and arithmetic intensity. Concretely, we observe the following trade-offs in our experiments:
\begin{itemize}
    \item \textbf{Row-major tiling:} improves locality for the input activation $X$, reducing repeated loads of input rows and benefiting scenarios with larger batch sizes or when activations dominate memory traffic.
    \item \textbf{Column-major tiling:} better reuses weight tiles, which is preferable when model weights dominate memory footprint (a common case in agentic decoding with small batch sizes and large models).
\end{itemize}


We integrate these tiling and loop-ordering choices into the fused kernel implementation and tune tile sizes to balance register usage, shared-memory occupancy, and Tensor Core utilization. Because the best kernel configuration depends on model shapes, batch size, GPU microarchitecture, and distributed interconnects, we accompany DeepFusionKernel with a lightweight profiler-driven scheduler. At deployment time, the scheduler quickly benchmarks the set of candidate kernels prior to the inference on the target hardware and selects the highest-throughput option for the given model and workload, ensuring robust performance across architectures and real-world inference conditions.

\section{Experiments}
\label{sec:experiments}

We integrate DeepFusionKernel into a complete inference pipeline and measure end-to-end decoding throughput in a realistic framework to isolate practical benefits and deployment considerations. To minimize kernel invocation and CPU overhead while maintaining a consistent evaluation environment, we run all end-to-end experiments inside the SGLang inference framework, with FlashInfer and CUDA Graphs enabled. We compare against three baselines: PyTorch (naive distributed), SGLang (default kernels), and vLLM. A lightweight kernel scheduler selects the best fused-kernel variant for each hardware/configuration before measurement.

\subsection{Experiment Setup}

Kernel performance depends on memory bandwidth, compute capability, model shape, and interconnect topology. To stress memory behavior, we run tensor-parallel inference across four NVIDIA A100 or H100 80 GB SXM GPUs (TP=4), which exposes both HBM access and inter-GPU communication effects. We first evaluate LLaMA 3.1 70B in FP16 with fixed prompt length 1 and target output lengths of 1024 tokens; batch sizes vary from 1 to 64. Each configuration is measured four times; we report mean throughput and standard deviation. For long-generation experiments (\Cref{sec:agentic}), output length is swept to probe KV-cache and attention effects.



\subsection{Full-Model Throughput}

\begin{table*}
\centering
\caption{Decoding throughput and standard deviation of Llama 3.1 70B with 1024 output length, using Torch, SGLang, and DeepFusionKernel. Throughput speedup rates of DeepFusionKernel against SGLang are presented.}
\label{tab:thp}
\resizebox{\linewidth}{!}{%
\begin{tabular}{ccccccccc}
\toprule
\multirow{3}{*}{\begin{tabular}[c]{@{}c@{}}Batch\\ size\end{tabular}} &
  \multicolumn{4}{c}{A100 GPU cluster} &
  \multicolumn{4}{c}{H100 GPU cluster} \\
\cmidrule(lr){2-5} \cmidrule(lr){6-9}
 &
  \multirow{2}{*}{\begin{tabular}[c]{@{}c@{}}Torch\\ Throughput\end{tabular}} &
  \multirow{2}{*}{\begin{tabular}[c]{@{}c@{}}SGLang\\ Throughput\end{tabular}} &
  \multicolumn{2}{c}{DeepFusionKernel} &
  \multirow{2}{*}{\begin{tabular}[c]{@{}c@{}}Torch\\ Throughput\end{tabular}} &
  \multirow{2}{*}{\begin{tabular}[c]{@{}c@{}}SGLang\\ Throughput\end{tabular}} &
  \multicolumn{2}{c}{DeepFusionKernel} \\
\cmidrule(lr){4-5} \cmidrule(lr){8-9} 
 &
   &
   &
  Throughput &
  Speedup &
   &
   &
  Throughput &
  Speedup \\
\midrule
1 &
  $2.97 \scriptstyle\pm 0.12$ &
  $35.64 \scriptstyle\pm 0.04$ &
  $37.66 \scriptstyle\pm 0.04$ &
  \textcolor{Green}{+5.7\%} &
  $3.55 \scriptstyle\pm 1.08$ &
  $60.2 \scriptstyle\pm 0.5$ &
  $62.22 \scriptstyle\pm 0.14$ &
  \textcolor{Green}{+3.4\%} \\
2 &
  $5.65 \scriptstyle\pm 0.41$ &
  $67.43 \scriptstyle\pm 2.64$ &
  $72.6 \scriptstyle\pm 0.04$ &
  \textcolor{Green}{+7.7\%} &
  $8.39 \scriptstyle\pm 1.4$ &
  $107.05 \scriptstyle\pm 18.21$ &
  $121.13 \scriptstyle\pm 0.57$ &
  \textcolor{Green}{+13.2\%} \\
4 &
  $11.83 \scriptstyle\pm 0.37$ &
  $134.02 \scriptstyle\pm 0.06$ &
  $140.55 \scriptstyle\pm 0.13$ &
  \textcolor{Green}{+4.9\%} &
  $18.11 \scriptstyle\pm 1.95$ &
  $228.9 \scriptstyle\pm 0.24$ &
  $237.05 \scriptstyle\pm 1.05$ &
  \textcolor{Green}{+3.6\%} \\
8 &
  $22.98 \scriptstyle\pm 1.67$ &
  $255.29 \scriptstyle\pm 0.09$ &
  $270.74 \scriptstyle\pm 0.46$ &
  \textcolor{Green}{+6.1\%} &
  $34.30 \scriptstyle\pm 4.31$ &
  $451.64 \scriptstyle\pm 0.72$ &
  $470.63 \scriptstyle\pm 0.33$ &
  \textcolor{Green}{+4.2\%} \\
16 &
  $40.57 \scriptstyle\pm 15.69$ &
  $453.75 \scriptstyle\pm 1.42$ &
  $497.69 \scriptstyle\pm 4.04$ &
  \textcolor{Green}{+9.7\%} &
  $52.37 \scriptstyle\pm 15.86$ &
  $878.08 \scriptstyle\pm 16.33$ &
  $914.62 \scriptstyle\pm 0.81$ &
  \textcolor{Green}{+4.2\%} \\
32 &
  $86.43 \scriptstyle\pm 17.27$ &
  $815.37 \scriptstyle\pm 0.4$ &
  $847.42 \scriptstyle\pm 2.55$ &
  \textcolor{Green}{+3.9\%} &
  $118.56 \scriptstyle\pm 13.25$ &
  $1551.33 \scriptstyle\pm 11.67$ &
  $1635.94 \scriptstyle\pm 11.39$ &
  \textcolor{Green}{+5.5\%} \\
64 &
  $169.33 \scriptstyle\pm 33.11$ &
  $1391.76 \scriptstyle\pm 1.97$ &
  $1410.39 \scriptstyle\pm 37.06$ &
  \textcolor{Green}{+1.3\%} &
  $129.81 \scriptstyle\pm 42.25$ &
  $3016.4 \scriptstyle\pm 9.74$ &
  $3119.55 \scriptstyle\pm 1.91$ &
  \textcolor{Green}{+3.4\%} \\
\bottomrule
\end{tabular}%
}
\vspace{5pt}
\end{table*}

\Cref{tab:thp} compares end-to-end decoding throughput across frameworks and GPUs. Consistent with prior work, SGLang substantially outperforms a naive distributed PyTorch baseline. Integrating DeepFusionKernel into SGLang yields additional gains--up to \textbf{9.7\%} on A100 and \textbf{13.2\%} on H100 for typical batch-size-limited decoding workloads--by reducing memory traffic in the SwiGLU MLPs.

The observed behavior follows expected bottleneck shifts: On A100, the benefit is largest at small batch sizes, where the workload is strongly memory-bandwidth-bound, and reduced reads/writes directly raise throughput. On H100, the kernel maintains an advantage across a wider batch-size range; H100's much higher compute throughput (1979 TFLOPs/s versus 312 TFLOPs/s on A100) means memory bandwidth continues to be a salient limiter, so memory-traffic reductions from fusion still translate to measurable speedups, though with diminishing marginal returns at very large batches.

Throughput variance grows with batch size across all setups, driven primarily by jitter in inter-GPU communication. Because DeepFusionKernel reuses SGLang's existing all-reduce and collective primitives, fusion does not change communication patterns and thus inherits this variance.




\subsection{Agentic Long-Generation Evaluation}
\label{sec:agentic}

Agentic workloads generate very long outputs and accumulate large KV caches, increasing off-chip memory pressure and per-token latency. To evaluate this regime, we benchmark LLaMA 3.1 70B for output lengths from 1024 up to 16384 tokens, using three representative batch sizes: $B=1$ (single-query), $B=4$ (light concurrent inference), and $B=16$ (moderate concurrency). Tests are run in FP16 and under TP=4 on both A100 and H100 clusters. For consistency, we measure only the decoding stage. 

Results in \Cref{tab:output-len} show DeepFusionKernel consistently improves throughput versus SGLang and vLLM across generation lengths and concurrency levels. Speedup variability is mostly explained by communication jitter and occasional system-level noise at high concurrency; nevertheless, the MLP remains a significant fraction of per-token latency even for long sequences, so memory-traffic reduction from fusion yields persistent gains.





\begin{table*}
\caption{Throughputs and standard deviation across batch sizes and output lengths using SGLang, vLLM, and DeepFusionKernel with kernel scheduler. Best throughputs are marked in bold.}
\label{tab:output-len}
\resizebox{\linewidth}{!}{%
\begin{tabular}{cccccccc}
\toprule
\multicolumn{2}{c}{GPU} &
  \multicolumn{3}{c}{A100} &
  \multicolumn{3}{c}{H100} \\
\cmidrule(lr){3-5} \cmidrule(lr){6-8}
Batch size &
  Output len &
  1024 &
  4096 &
  16384 &
  1024 &
  4096 &
  16384 \\
\midrule
\multirow{3}{*}{1} &
  SGLang &
  $35.68 \scriptstyle\pm 0.01$ &
  $35.53 \scriptstyle\pm 0.01$ &
  $34.97 \scriptstyle\pm 0.01$ &
  $60.20 \scriptstyle\pm 0.50$ &
  $60.77 \scriptstyle\pm 0.46$ &
  $60.18 \scriptstyle\pm 0.07$ \\
 &
  vLLM &
  $33.47 \scriptstyle\pm 0.21$ &
  $33.64 \scriptstyle\pm 0.05$ &
  $32.93 \scriptstyle\pm 0.51$ &
  $58.37 \scriptstyle\pm 0.12$ &
  $58.70 \scriptstyle\pm 0.05$ &
  $57.98 \scriptstyle\pm 0.12$ \\
 &
  DeepFusionKernel &
  $\mathbf{37.57 \scriptstyle\pm 0.10}$ &
  $\mathbf{37.50 \scriptstyle\pm 0.01}$ &
  $\mathbf{36.79 \scriptstyle\pm 0.05}$ &
  $\mathbf{61.22 \scriptstyle\pm 0.59}$ &
  $\mathbf{61.08 \scriptstyle\pm 0.31}$ &
  $\mathbf{60.72 \scriptstyle\pm 0.81}$ \\
\midrule
\multirow{3}{*}{4} &
  SGLang &
  $135.76 \scriptstyle\pm 0.11$ &
  $132.66 \scriptstyle\pm 0.44$ &
  $126.81 \scriptstyle\pm 0.16$ &
  $228.90 \scriptstyle\pm 0.24$ &
  $219.48 \scriptstyle\pm 3.74$ &
  $213.94 \scriptstyle\pm 1.03$ \\
 &
  vLLM &
  $126.10 \scriptstyle\pm 0.93$ &
  $125.10 \scriptstyle\pm 0.08$ &
  $119.30 \scriptstyle\pm 0.63$ &
  $227.69 \scriptstyle\pm 1.18$ &
  $227.70 \scriptstyle\pm 0.18$ &
  $218.73 \scriptstyle\pm 0.12$ \\
 &
  DeepFusionKernel &
  $\mathbf{140.15 \scriptstyle\pm 0.11}$ &
  $\mathbf{138.16 \scriptstyle\pm 0.02}$ &
  $\mathbf{131.75 \scriptstyle\pm 0.14}$ &
  $\mathbf{230.98 \scriptstyle\pm 1.79}$ &
  $\mathbf{234.61 \scriptstyle\pm 3.03}$ &
  $\mathbf{227.73 \scriptstyle\pm 0.04}$ \\
\midrule
\multirow{3}{*}{16} &
  SGLang &
  $453.75 \scriptstyle\pm 1.42$ &
  $452.60 \scriptstyle\pm 0.31$ &
  $396.70 \scriptstyle\pm 0.16$ &
  $878.08 \scriptstyle\pm 16.33$ &
  $815.40 \scriptstyle\pm 1.57$ &
  $732.18 \scriptstyle\pm 5.07$ \\
 &
  vLLM &
  $462.01 \scriptstyle\pm 1.56$ &
  $448.93 \scriptstyle\pm 0.43$ &
  $390.94 \scriptstyle\pm 0.63$ &
  $865.37 \scriptstyle\pm 3.98$ &
  $850.33 \scriptstyle\pm 0.47$ &
  $746.42 \scriptstyle\pm 0.47$ \\
 &
  DeepFusionKernel &
  $\mathbf{493.60 \scriptstyle\pm 2.75}$ &
  $\mathbf{478.66 \scriptstyle\pm 2.47}$ &
  $\mathbf{416.19 \scriptstyle\pm 0.79}$ &
  $\mathbf{894.32 \scriptstyle\pm 0.39}$ &
  $\mathbf{863.40 \scriptstyle\pm 0.13}$ &
  $\mathbf{764.57 \scriptstyle\pm 8.82}$ \\
\bottomrule
\end{tabular}%
}
\vspace{-10pt}
\end{table*}

\subsection{Discussion}

Our experimental results demonstrate that DeepFusionKernel with a profiler-driven scheduler consistently improves decoding throughput across realistic, bandwidth-bound inference scenarios. Key observations are:

\begin{itemize}\itemsep0em
    \item \textbf{Robust gains in memory-bound regimes.} Fusion produces the largest improvements when workloads are memory-bandwidth-limited (small batches, large models, long contexts). Even on H100, where compute is abundant, reducing memory traffic yields meaningful speedups.
    \item \textbf{Stable behavior for long-context generation.} As sequence length and KV cache grow, attention cost increases, but the MLP still contributes substantially to per-token cost; fusion therefore remains beneficial across long-generation workloads.
    \item \textbf{Low deployment overhead.} Kernel selection is performed with a short pre-inference profiling step; after selection, we use CUDA Graphs for capture, so there is no recurring inference-time dispatch overhead from the scheduler.
\end{itemize}

Overall, DeepFusionKernel provides a practical, deployable improvement to existing inference stacks: it reduces memory traffic in the critical SwiGLU path, integrates with current frameworks, and produces repeatable throughput gains across hardware and workload regimes.

\section{Related work}

Existing frameworks like Apex~\citep{Apex}, TensorRT-LLM~\citep{trt-llm}, and DeepSpeed-MII~\citep{ds-mii} use shallow fusion (e.g., GEMM+activation), leaving significant overheads for larger MLPs. 
Automatic compile-time fusion approaches exist: Welder~\citep{welder} fuses based on tile-graph cost models but is limited to linear chains; TVM~\citep{TVM} applies pattern matching and heuristics but its template-driven method mostly handles small trees; Blockbuster~\citep{blockbuster} uses algebraic rules and demonstrated a SwiGLU prototype, but remains a standalone compiler study and lacks runtime feedback and hardware-aware tuning.
In contrast, DeepFusionKernel deeply fuses the full SwiGLU into a single kernel. Paired with the kernel scheduler, it adapts fusion depth to workload and GPU, delivering consistent, branch-free speedups.

\section{Conclusion}
\label{sec:conclusion}

We present \textbf{DeepFusionKernel}, an aggressively fused CUDA operator that eliminates intermediate buffers in the SwiGLU MLP and rebalances the trade-off between memory traffic and on-chip compute. When integrated into SGLang and driven by a lightweight profiler-based scheduler, the fused kernel produces consistent, deployable throughput improvements--up to 9.7\% on A100 and 13.2\% on H100--across batch sizes and long-generation agentic workloads. By targeting the memory-bandwidth bottleneck that dominates autoregressive decoding, DeepFusionKernel lets modern GPUs better realize their available compute, making it a practical optimization for real-world LLM inference pipelines.

\section*{Limitations}

Due to limited computing resources, we do not exhaustively evaluate different GPU cluster interconnects, though our tests indicate that any resulting performance degradation is minimal for the workloads studied. Similarly, we do not quantify performance variation from inter-GPU communication, but its impact is mitigated by using the same inter-device reduction strategy as the baseline framework.


\bibliography{references}

\appendix
\setlength{\belowcaptionskip}{10pt}
\section{Technical Appendices and Supplementary Material}

\subsection{Distributed Inference}
\label{sec:distr-inference}

In distributed inference, computation is spread across multiple GPUs, either within a single server or across multiple server nodes. This setup introduces communication overheads, primarily due to all-reduce (for aggregating results) and all-gather (for collecting distributed outputs) operations. The frequency and cost of these operations are influenced by the kernel fusion strategy employed. The magnitude of the overheads generally scales with the volume of data transferred and can vary logarithmically, linearly, or quadratically with the number of devices or nodes, depending on the interconnect architecture.

A common strategy for distributed computation in large language models is \textbf{tensor parallelism (TP)}. For a matrix multiplication $Y = XW$, where $X \in \mathbb{R}^{M \times N}$, $W \in \mathbb{R}^{N \times K}$, and $Y \in \mathbb{R}^{M \times K}$, TP involves partitioning the weight matrix $W$ across devices, either row-wise (producing tall matrices) or column-wise (producing wide matrices). Each GPU computes a slice of the output $Y$, after which an all-gather operation is required to assemble the full result.

\begin{figure}
    \centering
    \includegraphics[width=\linewidth]{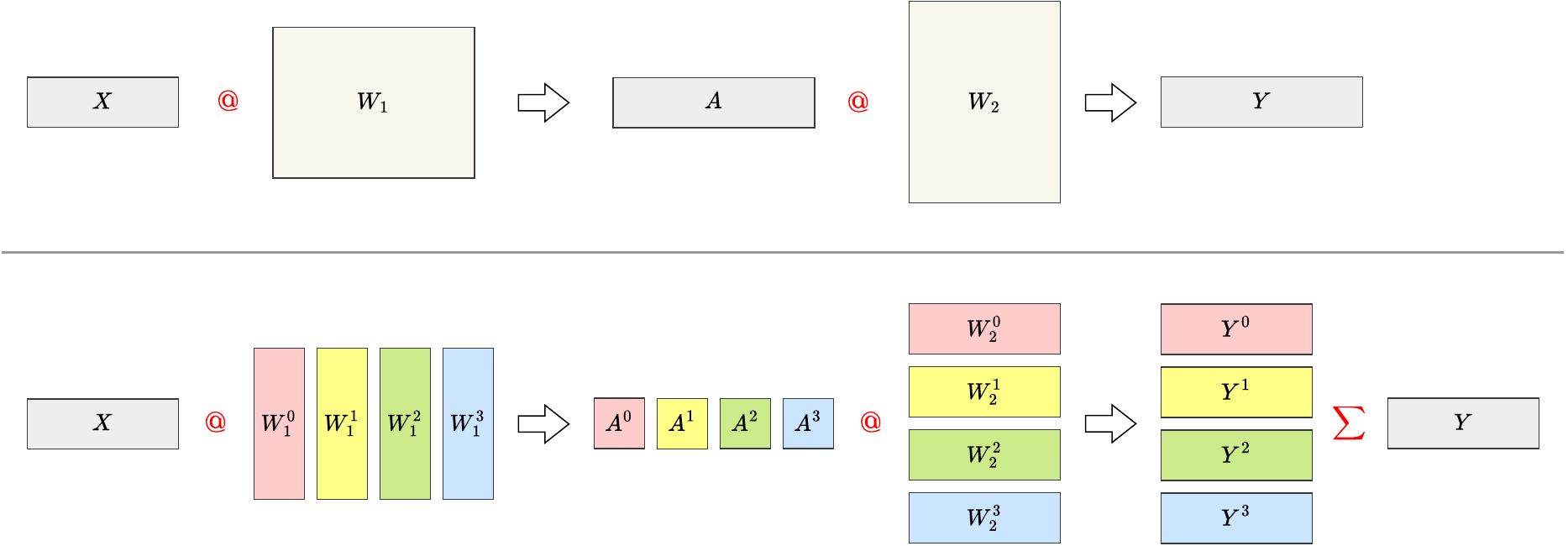}
    \caption{A matrix splitting scheme of consecutive matrix multiplications under tensor parallelism (TP) that contains a single all-reduce operation, denoted as \textcolor{red}{$\sum$}.}
    \label{fig:tp-matmul}
\end{figure}

However, in the case of compound matrix operations like SwiGLU MLP--which comprises two consecutive matrix multiplications with an intermediate nonlinearity--we can reduce communication overhead. Specifically, as demonstrated in \Cref{fig:tp-matmul}, we partition $W_\mathrm{Up}$ and $W_\mathrm{Gate}$ column-wise and $W_\mathrm{Down}$ row-wise. This configuration allows intermediate computations to remain local to each device, requiring only a single all-reduce at the end to aggregate the final output. Consequently, we reduce the number of collective operations to just one all-reduce per SwiGLU MLP block.

\subsection{Full-Model Performance Evaluation Experiment Setup}
\label{app:full-model-setup}

SGLang~\citep{sglang} is a high-performance inference framework for large language models (LLMs), demonstrating up to $6.4\times$ higher throughput than other state-of-the-art systems, including vLLM~\citep{vllm} and LMQL~\citep{LMQL}.

SGLang achieves this through a suite of custom optimizations tailored for efficient single- and multi-GPU inference, including:
\begin{itemize}
    \item \textbf{RadixAttention} for efficient key-value (KV) cache management;
    \item \textbf{Grouped GEMMs} for kernel fusion across parallel matrix multiplications;
    \item \textbf{Fused kernels} for common patterns of elementwise operations;
    \item \textbf{Optimized scheduling} for consecutive matrix multiplications under tensor parallelism (TP);
    \item \textbf{Custom cross-device all-reduce kernels};
    \item \textbf{TP-adapted model head}, combining a linear projection and a Softmax for next-token prediction.
\end{itemize}

SGLang supports several SOTA attention backends, including Triton~\citep{triton}, FlashAttention-3~\citep{flashattn3}, FlashMLA~\citep{flashmla}, and FlashInfer~\citep{flashinfer}. Additionally, SGLang captures CUDA Graphs during a warm-up phase, allowing inference to proceed without per-iteration PyTorch API or kernel launch overheads.

As kernel performance depends on both memory bandwidth and compute capacity, we evaluate DeepFusionKernel inference across multiple SOTA GPUs.

Due to the large model size, we use tensor parallelism (TP) to partition weight matrices across GPUs. TP does not increase the number of FLOPs or alter arithmetic intensity, assuming the fused computation strategy described in \Cref{sec:distr-inference} is followed.

However, TP introduces extra HBM access and communication latency during the all-reduce step that follows the TP-adapted $W_\mathrm{Down}$ projection. The impact of this latency grows with the amount of data to be synchronized and is sensitive to both the number of GPUs and their interconnect configuration. For example, in an NVIDIA A100 SXM cluster, each pair of GPUs is connected via NVIDIA NVLink (600GB/s), while the connections to the host system and between more GPUs employ the standard PCIe Gen4 links (64 GB/s)~\citep{a100}. Inter-device communication is also susceptible to packet drops, which introduces latency variability. 

For this evaluation, we use SGLang with the FlashInfer attention backend~\citep{flashinfer} and CUDA Graphs enabled so that once profiling identifies the best kernel, SGLang captures a complete CUDA Graph, eliminating branching during inference. We compare three setups:
\begin{itemize}
    \item Torch distributed inference
    \item SGLang with default kernels
    \item SGLang with DeepFusionKernel
\end{itemize}
We run Llama 3.1 70B in FP16 with TP across four NVIDIA A100 80GB SXM GPUs, all on a single node. To focus on decoding performance, we use an input sequence length of 1 and an output sequence length of 1024. Batch sizes range from $B=1$ to $B=64$. Each experiment is repeated four times, and we report the mean and standard deviation of decoding throughput.







\subsection{Licenses of Models and Frameworks}
\label{app:licenses}

\begin{itemize}
    \item Llama 3.1 70B~\citep{llama3}: Llama 3.1 Community License Agreement. Link: \url{https://huggingface.co/meta-llama/Llama-3.1-70B} and \url{https://huggingface.co/meta-llama/Llama-3.1-405B}.
    \item PyTorch 2.6.0~\citep{pytorch}: Please refer to the official GitHub page: \url{https://github.com/pytorch/pytorch/tree/v2.6.0}.
    \item SGLang v0.4.6.post4~\citep{sglang}: Apache-2.0 license. GitHub page link: \url{https://github.com/sgl-project/sglang/tree/v0.4.6.post4}.
    \item vLLM v0.10.1.1~\citep{vllm}: Apache-2.0 license. Github page link: \url{https://github.com/vllm-project/vllm/tree/v0.10.1.1}.
\end{itemize}










\end{document}